\documentclass{article}
\usepackage{ijcai17}
\usepackage{times}
\newcommand*\samethanks[1][\value{footnote}]{\footnotemark[#1]}
 
\title{Plan Explanations as Model Reconciliation:\\ Moving Beyond {\em Explanation as Soliloquy}}
 
\author{
Tathagata Chakraborti\thanks{Authors marked with asterix contributed equally.}\and Sarath Sreedharan\samethanks \and Yu Zhang \and Subbarao Kambhampati \\
School of Computing, Informatics, and Decision Systems Engineering\\
Arizona State University, Tempe, AZ 85281 USA\\
{ \{ tchakra2, ssreedh3, yzhan442, rao  \} @ asu.edu}
}
 
\usepackage[utf8]{inputenc}
\usepackage{graphicx}
\usepackage{amsmath}
\usepackage{amssymb}
\usepackage[dvipsnames]{xcolor}
 
 \usepackage[small]{caption}
 
\usepackage{algorithm}
\usepackage[noend]{algpseudocode}
\algrenewcommand\alglinenumber[1]{\tiny #1:}
 
\makeatletter
\def\BState{\State\hskip-\ALG@thistlm}
\makeatother

\usepackage{etoolbox}
\makeatletter
\patchcmd{\@verbatim}
  {\verbatim@font}
  {\verbatim@font\scriptsize}
  {}{}
\makeatother
 
\DeclareMathOperator*{\argmax}{arg\,max}
\DeclareMathOperator*{\argmin}{arg\,min}
 
\usepackage{amssymb}% http://ctan.org/pkg/amssymb
\usepackage{pifont}% http://ctan.org/pkg/pifont
\newcommand{\cmark}{\textcolor{black}{\ding{51}}}%
\newcommand{\xmark}{\textcolor{black}{\ding{55}}}%
\newcommand{\qmark}{\textcolor{black}{\textbf{?}}}%

\usepackage{booktabs}
\usepackage{multirow}
\usepackage{siunitx}
 
\begin{document}
 
\maketitle
 
\begin{abstract}
When AI systems interact with humans in the loop, they are often called on to provide explanations for their plans and behavior. Past work on plan explanations primarily involved the AI system explaining the correctness of its plan and the rationale for its decision in terms of its own model. Such soliloquy is wholly inadequate in most realistic scenarios where the humans have domain and task models that differ significantly from that used by the AI system. We posit that the explanations are best studied in light of these differing models. In particular, we show how explanation can be seen as a ``model reconciliation problem" (MRP), where the AI system in effect suggests changes to the human's model, so as to make its plan be optimal with respect to that changed human model. We will study the properties of such explanations, present algorithms for automatically computing them, and evaluate the performance of the algorithms. 
\end{abstract}
 
\section{Introduction}
There has been significant renewed interest recently in developing AI systems that can automatically provide explanations to humans in the loop. While much of the interest has been focused on learning systems that can explain their classification decisions, a related broader problem involves providing  explanations in the context of  human-AI  interaction and human-in-the-loop decision making systems. In such scenarios, the automated agents are called upon to provide explanation of their behavior or plans \cite{pat}. 
 
\begin{figure}[tpb!]
\centering
\includegraphics[width=0.95\columnwidth]{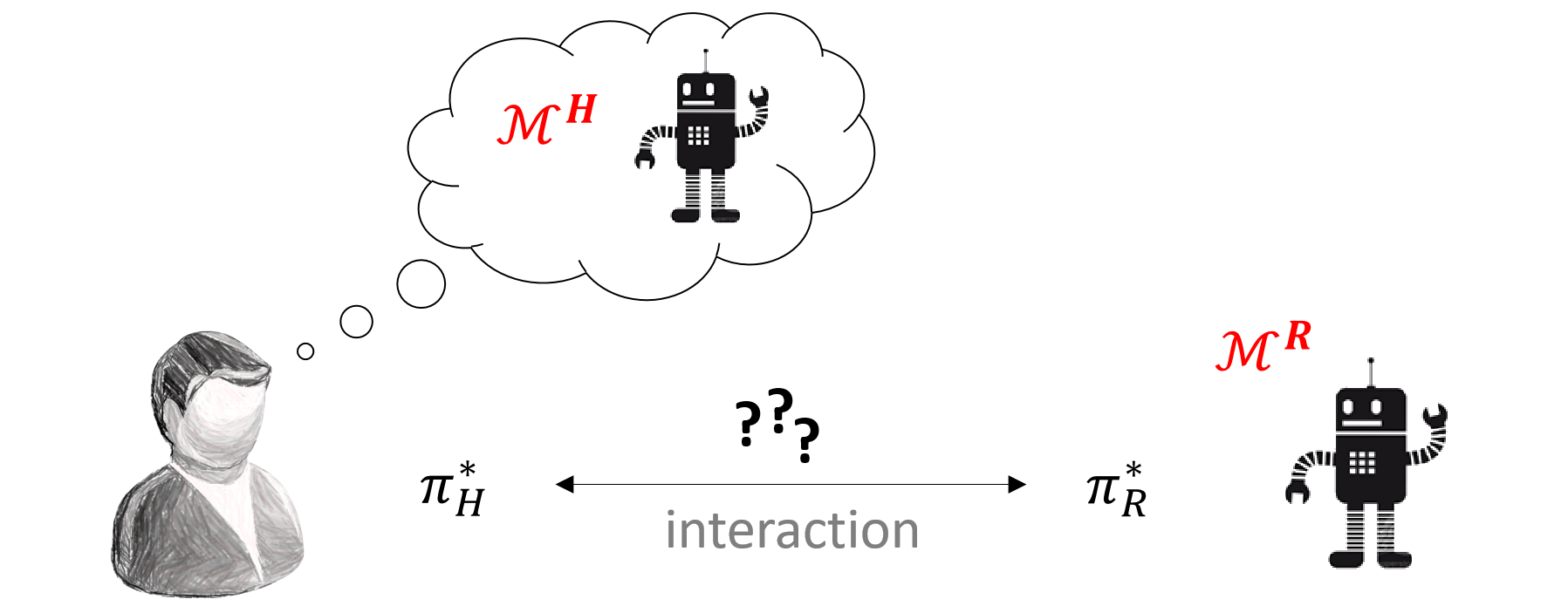}
\caption{
Interaction between humans and AI systems is best analyzed in light of their differing models. The robot here generates an optimal plan $\pi^*_R$ with respect to its model $\mathcal{M}^R$, which is interpreted by the human with respect to his model $\mathcal{M}^H$. Explanations are needed when $\pi^*_R$ is {\em not} an optimal plan with respect to $\mathcal{M}^H$. 
}
\vspace*{-0.05in}
\label{img}
\end{figure}
 
Although  explanation of plans has been investigated in the past (c.f. \cite{kambhampati1990classification,sohBiaMcIAAAI2011}), much of that work involved the planner explaining its decisions with respect to its own model (i.e. current state, actions and goals) and assuming that this ``{\em soliloquy}" also helps the human in the loop. While such a sanguine assumption may well be requited when the  human is an expert  ``debugger'' and is intimately familiar with the agent's innards, it is completely unrealistic in most human-AI interaction scenarios, where the humans may have a domain and task model that differs significantly from that used by the AI system. This is illustrated in Figure~\ref{img}, where the  plans generated by the AI system with respect to its model need to be interpreted by the human with respect to his model. Of course, the AI system can avoid the need to provide explanations by being ``{\em explicable}" \cite{exp-yu,exp-anagha} - i.e., generate plans that also make sense with respect to the humans' model. Such explicability requirement however puts additional constraints on the agent's plans, and may  not always be feasible. When the robot's plan is different from what the human would expect given his model of the world, the robot will be called on to ``explain" its plan. We posit that such explanations should be seen as the robot's attempt to move the human's model to be in conformance with its own. 
 
The primary contribution of this paper is to show how such model updates or \emph{explanations} can be formulated concisely as the \emph{model reconciliation problem} (MRP), which aims to make minimal changes to the human's model to bring it closer to the robot's model, in order to make the robot's plan optimal with respect to this changed human's model. 
One immediate complication in tackling an MRP is that the human's model is not directly made available to the robot, and will have to be learned 
instead (c.f. \cite{exp-yu}). 
The learned model may also be in a different form and at a different level of abstraction than the one used by the robot \cite{tian2016discovering,perera2016dynamic}. Nevertheless, for the purposes of this paper, we will assume that the human's model is made available and is in PDDL format, just like the robot's one. This allows us to focus on the explanation generation aspects. 
 
In the rest of the paper, we will formalize the scenario in Figure \ref{img} as the \emph{Multi-Model Planning} setting, and characterize explanation generation as a \emph{model reconciliation process} in it. 
We start by enumerating a few desirable requirements of such explanations - namely completeness, conciseness, monotonicity and computability. 
We then formulate different kinds of explanations that satisfy these requirements and relax one of these requirements at a time in the interests of computability. 
We present an A$^*$-search formulation for searching in the space of models to compute these explanations, and develop approximations and heuristics for the same.
Finally, we present a preliminary evaluation of the efficiency of our algorithms for generating explanations in randomly generated problems in a few benchmark planning domains. 

\begin{figure}
\centering
\includegraphics[width=\columnwidth]{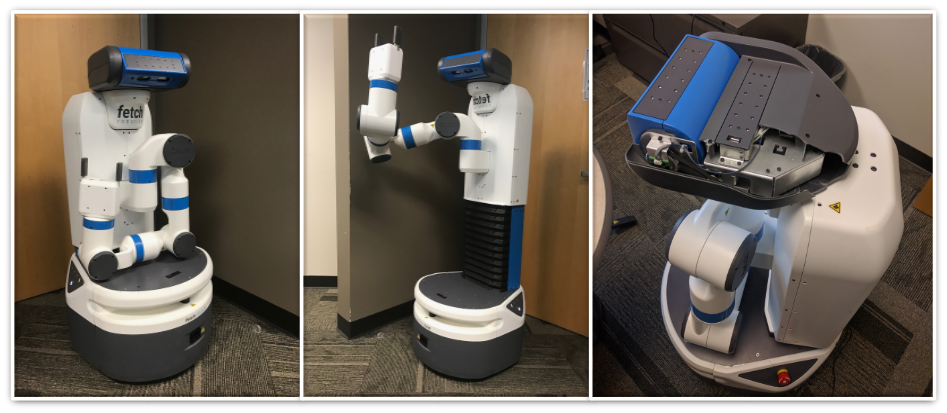}
\caption{The Fetch in the crouched position with arm tucked (left), torso raised and arm outstretched (middle) and the rather tragic consequences of a mistaken action model (right).
}
\vspace*{-0.05in}
\label{cheese}
\end{figure}
 
\vspace{-10pt}
\paragraph{A Motivating Example}
 
Let us illustrate the concept of explanations via model reconciliation through an example 
based on the Fetch robot whose design requires it to \texttt{tuck} its arms and lower its torso or \texttt{crouch} before moving - which is not obvious to a human navigating it. This may lead to an unbalanced base and toppling of the robot
if the human deems such actions as unnecessary. 
The move action for the robot is described in PDDL in the following model snippet -
 
\begin{verbatim}
(:action move
:parameters     (?from ?to - location)
:precondition   (and (robot-at ?from)
                     (hand-tucked) (crouched))
:effect         (and (robot-at ?to)
                     (not (robot-at ?from))))
(:action tuck
:parameters     ()
:precondition   ()
:effect         (and (hand-tucked)
                     (crouched)))
(:action crouch
:parameters     ()
:precondition   ()
:effect         (and (crouched)))
\end{verbatim}
 
Notice that the \texttt{tuck} action
also involves a lowering of torso so that the arm can rest on the base once it is tucked in. 
Now, consider a problem with the following initial and goal states (here, identical for both the robot and the human) -
 
\begin{verbatim}
(:init (block-at b1 loc1) (robot-at loc1) (hand-empty))  	          
(:goal (and (block-at b1 loc2)))
\end{verbatim}
 
An optimal plan for the robot, in this case, involves a \texttt{tuck} action followed by a \texttt{move} - 
 
\begin{verbatim}
pick-up b1 -> tuck -> move loc1 loc2 -> put-down b1
\end{verbatim}
 
The human, on the other hand, expects a much simpler model, as shown below. The \texttt{move} action does not have the preconditions for tucking the arm and lowering the torso, while \texttt{tuck} does not automatically lower the torso either.
 
\begin{verbatim}
(:action move
:parameters     (?from ?to - location)
:precondition   (and (robot-at ?from)
:effect         (and (robot-at ?to)
                     (not (robot-at ?from))))
(:action tuck
:parameters     ()
:precondition   ()
:effect         (and (hand-tucked))
 
(:action crouch
:parameters     ()
:precondition   ()
:effect         (and (crouched)))
\end{verbatim}
 
Clearly, the original plan is no longer optimal (and hence explicable) here. One possible model update (i.e. explanation) that can mitigate this situation is -
 
\begin{verbatim}
Explanation >> MOVE_LOC1_LOC2-has-precondition-HAND-TUCKED
\end{verbatim}
 
This correction brings the human and the robot model closer, and is necessary and sufficient to make the robot's plan optimal in the resultant domain. As indicated before, we refer to such model corrections as \emph{multi-model explanations}. 
 
\section{Related Work}
Our view of explanation as a model reconciliation  process is supported by studies in the field of psychology which stipulate that explanations \emph{``privilege a subset of beliefs, excluding possibilities inconsistent with those beliefs... can serve as a source of constraint in reasoning...''} \cite{Lombrozo2006464}. This is achieved in our case by the appropriate change in the expectation of the model that is believed to have engendered the plan in question. Further, authors in \cite{lombrozo2012explanation} also underline that explanations are \emph{``typically contrastive... the contrast
provides a constraint on what should figure in a selected
explanation...''} - this is especially relevant in order for an explanation to be self-contained and unambiguous. Hence the requirement of optimality in our explanations, which not only ensures that the current plan is valid in the updated model, but is also better than other alternatives. This is consistent with the notion of optimal (single-model) explanations investigated in \cite{sohBiaMcIAAAI2011} where less costly plans are referred to as \emph{preferred explanations}. 
The optimality criterion, however, makes the problem fundamentally different from model change algorithms in \cite{gobelbecker2010coming,Herzig:2014:RPT:3006652.3006726,siftmaintaining} which focus more on the feasibility of plans or correctness of domains. 

Finally, while the human-in-the-loop setting discussed here does bring back memories of mixed-initiative planners of the past \cite{ferguson1996trains,ai2004mapgen}, most of the work there involved the humans entering the land of planners; and not the other way around. Not surprisingly, it did not have the planner taking the human model into account in its planning or explanation. 
 
\section{The Multi-Model Planning (MMP) Setting}
 
\paragraph{A Classical Planning Problem}
\cite{russell2003artificial}
requires a model $\mathcal{M} = \langle \mathcal{D}, \mathcal{I}, \mathcal{G} \rangle$ 
(represented in PDDL \cite{mcdermott1998pddl}) 
consisting of the domain $\mathcal{D} = \langle F, A\rangle$ - where $F$ is a finite set of fluents that define the world state $s \subseteq F$, and $A$ is a finite set of actions - and the initial and goal states $\mathcal{I}, \mathcal{G} \subseteq F$. 
Action $a \in A$ is a tuple $\langle c_a, \textit{pre}(a), \textit{eff}^+(a), \textit{eff}^-(a)\rangle$ where $c_a$ denotes cost, and $\textit{pre}(a), \textit{eff}^+(a), \textit{eff}^-(a) \subseteq F$ is the set of preconditions and add / delete effects, i.e. $\delta_{\mathcal{M}}(s, a) \models \bot \textit{ if } s\not\models \textit{pre}(a); \textit{ else } \delta_{\mathcal{M}}(s, a) \models s \cup \textit{eff}^+(a) \setminus \textit{eff}^-(a)$ where $\delta_{\mathcal{M}}(\cdot)$ is the transition function. 
The cumulative transition function is given by $\delta_{\mathcal{M}}(s,\langle a_1, a_2, \ldots, a_n \rangle) = \delta_{\mathcal{M}}(\delta_{\mathcal{M}}(s, a_1),\langle a_2, \ldots, a_n \rangle)$.
The solution to the planning problem is a sequence of actions or a (satisficing) \emph{plan} $\pi = \langle a_1, a_2, \ldots, a_n \rangle$ such that $\delta_{\mathcal{M}}(\mathcal{I}, \pi) \models \mathcal{G}$. 
The cost of a plan $\pi$ is given by $C(\pi, \mathcal{M}) = \sum_{a\in\pi}c_a$ if $\delta_{\mathcal{M}}(\mathcal{I}, \pi) \models \mathcal{G}$; $\infty$ otherwise. 
The cheapest plan $\pi^* = \argmin_{\pi}\{C(\pi,\mathcal{M})~\forall \pi \text{ such that } \delta_{\mathcal{M}}(\mathcal{I}, \pi) \models \mathcal{G}\}$ is called the (cost) optimal plan. 
We will refer to the cost of the optimal plan in the model $\mathcal{M}$ as $C_\mathcal{M}^*$.
 
From the perspective of classical planning, the question of \textbf{plan explicability} may indeed be posed in terms of plan optimality. 
The intuition here is that if a plan is the best possible plan that the planner could have come up with, then it is also comprehensible to a human \emph{with the same planning model} and \emph{reasoning capabilities}, i.e. the existence of such a plan requires no further explanation.
However, the human's understanding of a planning problem often differs from the planner's, e.g. if she does not have access to the planner's actual goal, or does not know the current state accurately, or has a different perception of the action model being used.
In such situations, a plan $\pi$ produced in the robot's model $\mathcal{M}^R$ is being evaluated in terms of a different model $\mathcal{M}^H$ in the human's mind, as a result of which what is optimal (and explicable) in the planner's model may no longer be so in the human's, i.e $C(\pi, \mathcal{M}^R) = C_{\mathcal{M}^R}^*$, but $C(\pi, \mathcal{M}^H) > C_{\mathcal{M}^H}^*$.
Based on this, we define the following setting - 
 
\vspace{-10pt}
\paragraph{A Multi-Model Planning (MMP) Setting} is given by the tuple $\langle \mathcal{M}^R, \mathcal{M}^H \rangle$, where $\mathcal{M}^R = \langle D^R, \mathcal{I}^R, \mathcal{G}^R \rangle$ is the planner's model of the planning problem, while $\mathcal{M}^H = \langle D^H, \mathcal{I}^H, \mathcal{G}^H \rangle$ is the human's approximation of the same. 
 
\vspace{5pt}
As we mentioned in the introduction, from the point of view of the planner, there can be two approaches to achieve common ground with the human in such settings - \textbf{(1)} \emph{Change its own behavior in order to be explicable to the human} - in \cite{exp-yu,exp-anagha} the authors propose to modify the robot plan $\pi$ itself so that $C(\pi, \mathcal{M}^H) \approx C_{\mathcal{M}^H}^*~\wedge~\delta_{\mathcal{M}^R}(\mathcal{I}^R, \pi) \models \mathcal{G}^R$. Thus the planner chooses to sacrifice optimality
in order to make its behavior explicable to the human observer; and
\textbf{(2)} \emph{Bring the human's model closer to its own by means of explanations in the form of model updates} - here, the planner does not change its own behavior, but rather corrects the human's incorrect perception of its model via \emph{explanations}. 
We refer to this as \emph{the model reconciliation process}.
In this paper, we will focus on this only.
 
\vspace{-10pt}
\paragraph{The Model Reconciliation Problem (MRP)} 
is a tuple $\langle \pi^*, \langle \mathcal{M}^R, \mathcal{M}^H\rangle \rangle$ where $C(\pi^*, \mathcal{M}^R) = C_{\mathcal{M}^R}^*$, i.e. the robot's model, the human's approximation of it, and a plan that is optimal in the former.
 
\vspace{5pt}
Before we go into the details of the model reconciliation process, we will define the following state representation over planning problems. We intend to use this in our ``model-space search" for model reconciliation. 
{\scriptsize
\begin{align*}
\mathcal{F} =~& \{\textit{init-has-f}~|~\forall f \in F^H \cup F^R\} \cup\{\textit{goal-has-f}~|~ \forall f \in F^H \cup F^R\}\\
& \bigcup_{a \in A^H \cup A^R}\{\textit{a-has-precondition-f}, \textit{a-has-add-effect-f}, \\[-3ex]
& ~~~~~~~~~~~~~~~~~~~~~~~\textit{a-has-del-effect-f}~|~ \forall f \in F^H \cup F^R\}\\
& \cup \{\textit{a-has-cost-}c_a~|~a \in A^H\} \cup \{\textit{a-has-cost-}c_a~|~a \in A^R\}.
\end{align*}
}%
A mapping function $\Gamma : \mathcal{M} \mapsto s$ represents any planning problem $\mathcal{M} = \langle \langle F, A\rangle, \mathcal{I}, \mathcal{G}\rangle$ as a state $s \subseteq \mathcal{F}$ as follows -
{\scriptsize
\begin{align*}
\label{eqn2}
    \tau(f) &= 
\begin{cases}
    \textit{init-has-f} & \text{ if } f \in \mathcal{I},\\
    \textit{goal-has-f} & \text{ if } f \in \mathcal{G},\\
    \textit{a-has-precondition-f} & \text{ if } f \in \textit{pre}(a),~a \in A\\
    \textit{a-has-add-effect-f} & \text{ if } f \in \textit{eff}^+(a),~a \in A\\
    \textit{a-has-del-effect-f} & \text{ if } f \in \textit{eff}^-(a),~a \in A\\
    \textit{a-has-cost-f} & \text{ if } f = c_a,~a \in A\\
\end{cases} \nonumber \\[1ex]
	\Gamma(\mathcal{M}) =~& \big\{\tau(f)~|~\forall f \in \mathcal{I} \cup \mathcal{G} \cup \nonumber \\ 
    & \bigcup_{a\in A}\{f'~|~\forall f' \in \{c_a\} \cup \textit{pre}(a) \cup \textit{eff}^+(a) \cup \textit{eff}^-(a)\}\big\}
\end{align*}
}%
We can now define a \emph{model-space search problem} $\langle \langle \mathcal{F}, \Lambda \rangle, \Gamma(\mathcal{M}_1), \Gamma(\mathcal{M}_2)\rangle$ 
with a new action set $\Lambda$ containing unit model change actions $\lambda : \mathcal{F} \rightarrow \mathcal{F}$ such that $| s_1 \Delta s_2 | = 1$, where the new transition or edit function is given by $\delta_{\mathcal{M}_1,\mathcal{M}_2}(s_1, \lambda) = s_2$ such that \texttt{condition 1} : $s_2 \setminus s_1 \subseteq \Gamma(\mathcal{M}_2)$ and \texttt{condition 2} : $s_1 \setminus s_2 \not\subseteq \Gamma(\mathcal{M}_2)$ are satisfied. 
This means that model change actions can only make a single change to a domain at a time, and \emph{all these changes are consistent
with the model of the planner. }
The solution to a model-space search problem is given by a set of edit functions $\{\lambda_i\}$ that can transform the model $\mathcal{M}_1$ to the model $\mathcal{M}_2$, i.e. $\delta_{\mathcal{M}_1,\mathcal{M}_2}(\Gamma(\mathcal{M}_1), \{\lambda_i\}) = \Gamma(\mathcal{M}_2)$.
 
\vspace{-10pt}
\paragraph{A Multi-Model Explanation} 
denoted by $\mathcal{E}$ is a solution to an MRP, i.e. a solution to the model-space search problem $\langle \langle \mathcal{F}, \Lambda \rangle, \Gamma(\mathcal{M}^H), \Gamma(\widehat{\mathcal{M}})\rangle$
with the transition function $\delta_{\mathcal{M}_H, \mathcal{M}_R}$
such that $C(\pi^*, \widehat{\mathcal{M}}) - C_{\widehat{\mathcal{M}}}^* < C(\pi^*, \mathcal{M}^H) - C_{\mathcal{M}^H}^*$. 
 
\vspace{5pt}
This means that in the updated model after the explanation, the plan in question is closer to the optimal (and hence less inexplicable) than it was in the original model that the human had. The human in the loop can either chose to use this explanation to update her own model or negotiate in course of further dialog (e.g. to update the robot's model). 
 
As we go on to develop approaches to compute different types of such explanations, we will consider the following four requirements that characterize each solution.
 
\begin{itemize}
\item[R1.] \textbf{Completeness -} Explanations of a plan should be able to be compared and contrasted against other alternatives, so that no better solution exists. We enforce this property by requiring that in the updated human model the plan being explained is optimal.
\begin{itemize}
\item {\em An explanation is complete iff $C(\pi^*, \widehat{\mathcal{M}}) = C_{\widehat{\mathcal{M}}}^*$.}
\end{itemize}
\item[R2.] \textbf{Conciseness -} 
Explanation should be concise so that they are easily understandable to the explainee. Larger an explanation is, the harder it is for the human to incorporate that information into her deliberative process.\item[R3.] \textbf{Monotonicity -} This ensures that remaining model differences cannot change the completeness of an explanation, i.e. all aspects of the model that engendered the plan have been reconciled. 
This thus subsumes completeness and requires more detailed\footnote{This is a very useful property to have. Doctors, for example, reveal different amount of details of their model to their patients as opposed to their peers. Further, the idea of completeness, i.e. withholding information on other model changes as long as they explain the observed plan, is also quite prevalent in how we deal with similar scenarios ourselves - e.g. progressing from Newtonian physics in high school to Einsteins Laws of Relativity in college.} explanations.
\begin{itemize}
\item {\em An explanation is monotonic iff $C(\pi^*, \hat{\mathcal{M}}) = C^{*}_{\hat{\mathcal{M}}}$ $\forall\hat{\mathcal{M}} : \Gamma(\widehat{\mathcal{M}}) \Delta \Gamma(\mathcal{M}^H) \subset \Gamma(\hat{\mathcal{M}}) \Delta \Gamma(\mathcal{M}^H)$.}
\end{itemize}
\item[R4.] \textbf{Computability -} While conciseness deals with how easy it is for the explainee to understand an explanation, computability measures the ease of computing the explanation from the point of view of the planner.
\end{itemize}
 
We will now introduce different kinds of multi-model explanations that can participate in the model reconciliation process, propose algorithms to compute them, and compare and contrast their respective properties. 
We note that the requirements outlined above are in fact often at odds with each other - an explanation that is very easy to compute may be very hard to comprehend. This (as seen in Table \ref{table-vs}) will become clearer in course of this discussion. 
 
A simple way to compute an explanation would be to provide the model differences pertaining to only the actions that are present in the plan that needs to be explained - 
 
\vspace{-10pt}
\paragraph{A Plan Patch Explanation (PPE)} is given by $\mathcal{E}^{MPE} = \Delta_{i \in \{H,R\}}\bigcup_{f \in \{c_a\} \cup \textit{pre}(a) \cup \textit{eff}^+(a) \cup \textit{eff}^-(a)~:~a \in \pi^* \cap A^i} \tau(f)$.
 
\vspace{5pt}
Clearly, such an explanation is easy to compute and concise by focusing only on plan being explained. However, it may also contain information that need not have been revealed, while at the same time ignoring model differences elsewhere in $\mathcal{M}^H$ that could have contributed to the plan being suboptimal in it. Thus, it is incomplete. An adoption of VAL \cite{howey2004val} to the MMP setting will be, in fact, a subset of such PPEs, and suffer from the same limitations. An easy way to compute a complete explanation would be to provide the entire model difference to the human --

\vspace{-10pt}
\paragraph{A Model Patch Explanation (MPE)} is given by $\mathcal{E}^{MPE} =  \Gamma(\mathcal{M}^R)\Delta\Gamma(\mathcal{M}^H)$.
 
\vspace{5pt}
This is clearly also easy to compute but can be quite large and is hence far from being concise. 
Thus, in the rest of paper, we will try to minimize the size (and hence increase the comprehensibility) of explanations by searching in the space of models and thereby not exposing information that is not relevant to the plan being explained while still trying to satisfy as many requirements as we can. 

%Keeping this in mind, we define --

%\subsubsection{Complete Explanations}

\vspace{-10pt}
\paragraph{A Minimally Complete Explanation (MCE)} is the shortest complete explanation, i.e.
\begin{itemize}
\item[] $\mathcal{E}^{MCE} = \argmin_{\mathcal{E}}| \Gamma(\widehat{\mathcal{M}}) \Delta \Gamma(\mathcal{M}^H) |$ with R1.
\end{itemize}
 
%\vspace{5pt}
The explanation provided before in the Fetch domain is indeed the smallest domain change that may be made to make the given plan optimal in the updated action model, and is thus an example of a minimally complete explanation.
An implicit assumptions we make here is that the computation power (or planning capability) of the human is is the same as that of the planner. This means that the human can compute the optimal plan given a planning problem. 

%The strictness of this assumption can be relaxed by requiring $\mathcal{E}^H(\pi^*, \widehat{\mathcal{M}}) > \epsilon,~\epsilon \in [0,1]$ in the solution of an MRP, to model an $\epsilon$-optimal human, or look at top-K plans \cite{RiaSohUdrSPARK14} for hypothesis generation.
%Of course, results on such a model will require validation through human-factors experiments, which is out of scope of this paper.

\begin{table}[tbp!]
\tiny
\begin{center}
    \begin{tabular}{| c || c | c | c | c |}
    \hline
    {\bf Explanation Type} & {\bf R1} & {\bf R2}  & {\bf R3}  & {\bf R4} \\ \hline \hline
    Plan Patch Explanation  / VAL                  & \xmark & \cmark & \xmark & \cmark \\\hline
    Model Patch Explanation                        & \cmark & \xmark & \cmark & \cmark \\\hline
    Minimally Complete Explanation                 & \cmark & \cmark & \xmark & \qmark \\\hline
    Minimally Monotonic Explanation                & \cmark & \cmark & \cmark & \qmark \\\hline
    (Approximate) Minimally Complete Explanation   & \xmark & \cmark & \xmark & \cmark \\
    \hline \hline
    \end{tabular}
\end{center}
\caption{
Requirements for different types of explanations.}
\label{table-vs}
\end{table}

\vspace{-10pt}
\paragraph{Model-space search for MCEs}
 
To compute MCEs, we employ A$^*$ search, similar to \cite{wayllacegoal}, in the space of models, as shown in Algorithm \ref{algo1}. 
Given an MRP, we start off with the initial state $\Gamma(\mathcal{M}^H)$ derived from the human's expectation of a given planning problem $\mathcal{M}^R$, and modify it incrementally until we arrive at a planning problem $\widehat{\mathcal{M}}$ with $C(\pi^*, \widehat{\mathcal{M}}) = C^{*}_{\widehat{\mathcal{M}}}$, i.e. the given plan is explained. 
Note that the model changes are represented as a set, i.e. there is no sequentiality in the search problem. Also, we assign equal importance to all model corrections. We can easily capture differential importance of model updates by attaching costs to the edit actions $\lambda$ - the algorithm remains unchanged.
 
We also employ a selection strategy of successor nodes to speed up search (by overloading the way the priority queue is being popped) by first processing model changes that are relevant to the actions in $\pi^*_R$ and $\pi_H$ before the rest. %More on this later in the evaluations section.
 
\vspace{-10pt}
\paragraph{Proposition 1} The successor selection strategy outlined in Algorithm \ref{algo1} yields an admissible heuristic for model space search for minimally complete explanations.
 
\vspace{-10pt}
\paragraph{Proof}
 
%At any point during MCE-SEARCH, 
%For an MCE $\mathcal{E}' \subset \mathcal{E}$, the set $\mathcal{E} \setminus \mathcal{E}'$ must contain at least one $\lambda$ related to actions in the set $ \{a~|~a \in \pi^*_{R} \wedge a \in \pi'\}$ (where $C(\pi',\delta(~\Gamma(\mathcal{M}^R),\mathcal{E}')) = C^{*}_{\delta(~\Gamma(\mathcal{M}^R),\mathcal{E}')}$). 
Let $\mathcal{E}$ be the MCE for an MRP problem and let $\mathcal{E}'$ be any intermediate explanation found by our search such that $\mathcal{E}' \subset \mathcal{E}$, then the set $\mathcal{E} \setminus \mathcal{E}'$ must contain at least one $\lambda$ related to actions in the set $ \{a~|~a \in \pi^*_{R} \vee a \in \pi'\}$ (where $\pi'$ is the optimal plan for the model $\hat{\mathcal{M}}$ where $\delta_{\mathcal{M}_H,\mathcal{M}_R}(\Gamma(\mathcal{M}^H),\mathcal{E}') = \Gamma(\hat{\mathcal{M}}$). 
To see why this is true, consider an $\mathcal{E}'$ where $|\mathcal{E}'| = |\mathcal{E}| - 1$. If the action in  $\mathcal{E} \setminus \mathcal{E}'$ does not belong to either $\pi^*_R$ or $\pi'$ then it can not improve the cost of $\pi^*_R$ in comparison to $\pi'$ and hence $\mathcal{E}$ can not be the MCE. 
Similarly we can show that this relation will hold for any size of $\mathcal{E}'$. We can leverage this knowledge about  $\mathcal{E} \setminus \mathcal{E}'$ to create an admissible heuristic that will only consider the relevant changes at any given point of time (by giving very large values to all other changes).
 
\vspace{5pt}
We also note that the optimality criterion is relevant to both the cases where the human expectation is better, or when it is worse, than the plan computed by the planner. This might be counter to intuition, since in the latter case one might expect that just establishing feasibility of a better (than expected optimal) plan would be enough. Unfortunately, this is not the case, as can be easily seen by creating counter-examples where other faulty parts of the human model might disprove the optimality of the plan in the new model -- 
 
\vspace{-10pt}
\paragraph{Proposition 2} \label{prop2}
 If $C(\pi^*, \mathcal{M}^H) < \min_{\pi} C(\pi, \mathcal{M}^H)$, then ensuring feasibility of the plan in the modified planning problem, i.e. $\delta_{\mathcal{\widehat{M}}}(\mathcal{\widehat{I}}, \pi^*) \models \mathcal{\widehat{G}}$, is a necessary but \emph{not} a sufficient condition for $\widehat{\mathcal{M}} = \langle \widehat{D}, \mathcal{\widehat{I}}, \mathcal{\widehat{G}} \rangle$ to yield a valid explanation.
 
\vspace{5pt}
Note that a minimally complete explanation for an MRP can be rendered invalid given further updates to the model. This can be easily demonstrated in our running example in the Fetch domain. Imagine that if, at some point, the human were to find out that the action move also has a precondition \texttt{(crouched)}, then the previous robot plan will no longer make sense to the human since now, according to the human's faulty model (being unaware that the tucking action also lowers the robot's torso) the robot would need to do \emph{both} \texttt{tuck} and \texttt{crouch} actions before moving.
Consider the following explanation in the Fetch domain instead --
 
\begin{verbatim}
Explanation >> 
TUCK-has-add-effect-CROUCHED
MOVE_LOC2_LOC1-has-precondition-CROUCHED
\end{verbatim}
 
This explanation does not reveal all model differences but at the same time ensures that the robot's plan remains optimal for this problem, irrespective of any other changes to the model, by accounting for all the relevant parts of the model that engendered the plan. It is also the smallest possible among all such explanations. 
The requirement of monotonicity and minimality brings us to the notion of -

\begin{algorithm}[tbp!]
\scriptsize
\caption{Search for Minimally Complete Explanations}
\label{algo1}
\begin{algorithmic}[1]
\Procedure{MCE-Search}{}
\vspace{2pt} 
\BState \emph{Input}:~~~~MRP $\langle \pi^*, \langle \mathcal{M}^R, \mathcal{M}^H\rangle\rangle$
\BState \emph{Output}: Explanation $\mathcal{E}^{MCE}$
\vspace{2pt} 
\BState \emph{Procedure}:  
\vspace{2pt} 
\State fringe~~~~~~~~$\leftarrow$ \texttt{Priority\_Queue()}
\State c\_list~~~~~~~~~~$\leftarrow$ \{\}  \Comment{\textcolor{black}{Closed list}}
\State $\pi^*_R$~~~~~~~~~~~~$\leftarrow \pi^*$ \Comment{\textcolor{black}{Optimal plan being explained}}
\State $\pi_H$~~~~~~~~~~~~$\leftarrow \pi$ such that $C(\pi, \mathcal{M}^H) = C^{*}_{\mathcal{M}^H}$ \Comment{\textcolor{black}{Plan expected by human}}
\State $\text{fringe.push}(\langle \mathcal{M}^H, \{\}\rangle,~\text{priority} = 0)$
\vspace{2pt} 
\While{True}
\vspace{2pt} 
\State $\langle \widehat{\mathcal{M}}, \mathcal{E} \rangle, c \leftarrow \text{fringe.pop}(\widehat{\mathcal{M}})$
\If{$C(\pi^*_R, \widehat{\mathcal{M}}) = C^{*}_{\widehat{\mathcal{M}}}$} return $\mathcal{E}$  \Comment{\textcolor{black}{Return $\mathcal{E}$ if $\pi^*_R$ optimal in   $\widehat{\mathcal{M}}$}}
\Else
\State c\_list $\leftarrow$ c\_list $\cup~\widehat{\mathcal{M}}$
\vspace{2pt} 
\For{$f \in \Gamma(\widehat{\mathcal{M}})~\setminus~\Gamma(\mathcal{M}^R)$} \Comment{\textcolor{black}{Models that satisfy condition 1}}
\State $\lambda \leftarrow \langle 1, \{\widehat{\mathcal{M}}\}, \{\}, \{f\} \rangle$ \Comment{\textcolor{black}{Removes f from $\widehat{\mathcal{M}}$}}
\If{$\delta_{\mathcal{M}^H,\mathcal{M}^R}(\Gamma(\widehat{\mathcal{M}}), \lambda) \not\in \text{c\_list}$}
\State $\text{fringe.push}(\langle \delta_{\mathcal{M}^H,\mathcal{M}^R}(\Gamma(\widehat{\mathcal{M}}), \lambda),~\mathcal{E}~\cup~\lambda \rangle,~c + 1)$
\EndIf
\EndFor
\vspace{2pt} 
\For{$f \in \Gamma(\mathcal{M}^R)~\setminus~\Gamma(\widehat{\mathcal{M}})$} \Comment{\textcolor{black}{Models that satisfy condition 2}}
\State $\lambda \leftarrow \langle 1, \{\widehat{\mathcal{M}}\}, \{f\}, \{\} \rangle$ \Comment{\textcolor{black}{Adds f to $\widehat{\mathcal{M}}$}}
\If{$\delta_{\mathcal{M}^H,\mathcal{M}^R}(\Gamma(\widehat{\mathcal{M}}), \lambda) \not\in \text{c\_list}$}
\State $\text{fringe.push}(\langle \delta_{\mathcal{M}^H,\mathcal{M}^R}(\Gamma(\widehat{\mathcal{M}}), \lambda),~\mathcal{E}~\cup~\lambda \rangle,~c + 1)$
\EndIf
\EndFor
\EndIf
\EndWhile
\EndProcedure
\vspace{4pt} 
\Procedure{Priority\_Queue.pop}{$\hat{\mathcal{M}}$}
\vspace{2pt} 
\State $\text{candidates} \leftarrow \{\langle \langle \widehat{\mathcal{M}}, \mathcal{E} \rangle, c^*\rangle ~|~ c^* = \argmin_{c}\langle \langle \widehat{\mathcal{M}}, \mathcal{E} \rangle, c\rangle\}$
\State $\text{pruned\_list} \leftarrow \{\}$
\State $\pi_H $~~~~~~~~~~~~~$ \leftarrow \pi$ such that $C(\pi, \hat{\mathcal{M}}) = C^{*}_{\hat{\mathcal{M}}}$
\For{$\langle \langle \widehat{\mathcal{M}}, \mathcal{E} \rangle, c\rangle \in$ candidates} 
\If{$\exists a \in \pi^*_R \cup \pi_H \text{ such that } \tau^{-1}(\Gamma(\widehat{\mathcal{M}})~\Delta~\Gamma(\hat{\mathcal{M}})) \in \{c_a\} \cup \textit{pre}(a) \cup \textit{eff}^+(a) \cup \textit{eff}^-(a)$} \Comment{\textcolor{black}{ Candidates relevant to $\pi^*_R$ or $\pi_H$}}
\State 
$\text{pruned\_list} \leftarrow \text{pruned\_list}~\cup~\langle \langle \widehat{\mathcal{M}}, \mathcal{E} \rangle, c\rangle$
\EndIf
\EndFor
\vspace{3pt} 
\If{$\text{pruned\_list} = \phi$}
$\langle \widehat{\mathcal{M}}, \mathcal{E} \rangle, c \sim Unif(\text{candidate\_list})$
\Else
~~~~~~~~~~~~~~~~~~~~~~~~~~~~~~~~~$\langle \widehat{\mathcal{M}}, \mathcal{E} \rangle, c \sim Unif(\text{pruned\_list})$
\EndIf 
\EndProcedure
\end{algorithmic}
\end{algorithm}
 
\begin{algorithm}[tbp!]
\scriptsize
\caption{Search for Minimally Monotonic Explanations}
\label{algo2}
\begin{algorithmic}[1]
  \Procedure{MME-Search}{}
\vspace{2pt} 
\BState \emph{Input}: MRP $\langle \pi^*, \langle \mathcal{M}^R, \mathcal{M}^H\rangle\rangle$
\BState \emph{Output}: Explanation $\mathcal{E}^{MME}$
\vspace{2pt} 
\BState \emph{Procedure}:  
\vspace{2pt} 
\State $\mathcal{E}^{MME}$~~~~$\leftarrow$ \{\}
\State fringe~~~~~~~~$\leftarrow$ \texttt{Priority\_Queue()}
\State c\_list~~~~~~~~~$\leftarrow$ \{\} \Comment{\textcolor{black}{Closed list}}
\State h\_list~~~~~~~~~$\leftarrow$ \{\} \Comment{\textcolor{black}{List of incorrect model changes}}
\State $\text{fringe.push}(\langle \mathcal{M}^R, \{\}\rangle,~\text{priority} = 0)$
\vspace{2pt} 
\While{fringe is not empty}
\vspace{2pt} 
\State $\langle \widehat{\mathcal{M}}, \mathcal{E} \rangle, c \leftarrow \text{fringe.pop}(\widehat{\mathcal{M}})$
\If{$C(\pi^*, \widehat{\mathcal{M}}) > C^{*}_{\widehat{\mathcal{M}}}$}
\State $\text{h\_list} \leftarrow \text{h\_list}~\cup~(\Gamma(\widehat{\mathcal{M}})~\Delta~\Gamma(\mathcal{M}^R))$ \Comment{\textcolor{black}{Updating h\_list }}
\Else
\State c\_list $\leftarrow$ c\_list $\cup~\widehat{\mathcal{M}}$
\vspace{2pt} 
\For{$f \in \Gamma(\widehat{\mathcal{M}})~\setminus~\Gamma(\mathcal{M}^H)$}  \Comment{\textcolor{black}{Models that satisfy condition 1}}
\State $\lambda \leftarrow \langle 1, \{\widehat{\mathcal{M}}\}, \{\}, \{f\} \rangle$ \Comment{\textcolor{black}{Removes f from $\widehat{\mathcal{M}}$}}
\If{$\delta_{\mathcal{M}^R,\mathcal{M}^H}(\Gamma(\widehat{\mathcal{M}}), \lambda) \not\in \text{c\_list} \newline \indent \indent \indent \textbf{~~~~and } \nexists S \text{ s.t. }(\Gamma(\widehat{\mathcal{M}})\Delta\Gamma(\mathcal{M}^R)) \supseteq S \in \text{h\_list}$} \Comment{\textcolor{black}{Prop 3}}
\State $\text{fringe.push}(\langle \delta_{\mathcal{M}^R,\mathcal{M}^H}(\Gamma(\widehat{\mathcal{M}}), \lambda),~\mathcal{E}~\cup~\lambda \rangle,~c + 1)$
\State $\mathcal{E}^{MME} \leftarrow \max_{|\cdot|}\{\mathcal{E}^{MME}, \mathcal{E}\}$
\EndIf
\EndFor
\vspace{2pt} 
\For{$f \in \Gamma(\mathcal{M}^H)~\setminus~\Gamma(\widehat{\mathcal{M}})$}  \Comment{\textcolor{black}{Models that satisfy condition 2}}
\State $\lambda \leftarrow \langle 1, \{\widehat{\mathcal{M}}\}, \{f\}, \{\} \rangle$ \Comment{\textcolor{black}{Adds f from $\widehat{\mathcal{M}}$}}
\If{$\delta_{\mathcal{M}^R,\mathcal{M}^H}(\Gamma(\widehat{\mathcal{M}}), \lambda) \not\in \text{c\_list} \newline \indent \indent \indent \textbf{~~~~and } \nexists S \text{ s.t. }(\Gamma(\widehat{\mathcal{M}})\Delta\Gamma(\mathcal{M}^R)) \supseteq S \in \text{h\_list}$} \Comment{\textcolor{black}{Prop 3}}
\State $\text{fringe.push}(\langle \delta_{\mathcal{M}^R,\mathcal{M}^H}(\Gamma(\widehat{\mathcal{M}}), \lambda),~\mathcal{E}~\cup~\lambda \rangle,~c + 1)$
\State $\mathcal{E}^{MME} \leftarrow \max_{|\cdot|}\{\mathcal{E}^{MME}, \mathcal{E}\}$
\EndIf
\EndFor
\EndIf
\EndWhile
\vspace{2pt} 
\State $\mathcal{E}^{MME} \leftarrow  (\Gamma(\widehat{\mathcal{M}})~\Delta~\Gamma(\mathcal{M}^R)) \setminus \mathcal{E}^{MME} $
\State return $\mathcal{E}^{MME}$
\vspace{2pt} 
\EndProcedure
\end{algorithmic}
\end{algorithm}

\begin{figure*}
\centering
\includegraphics[width=0.9\textwidth]{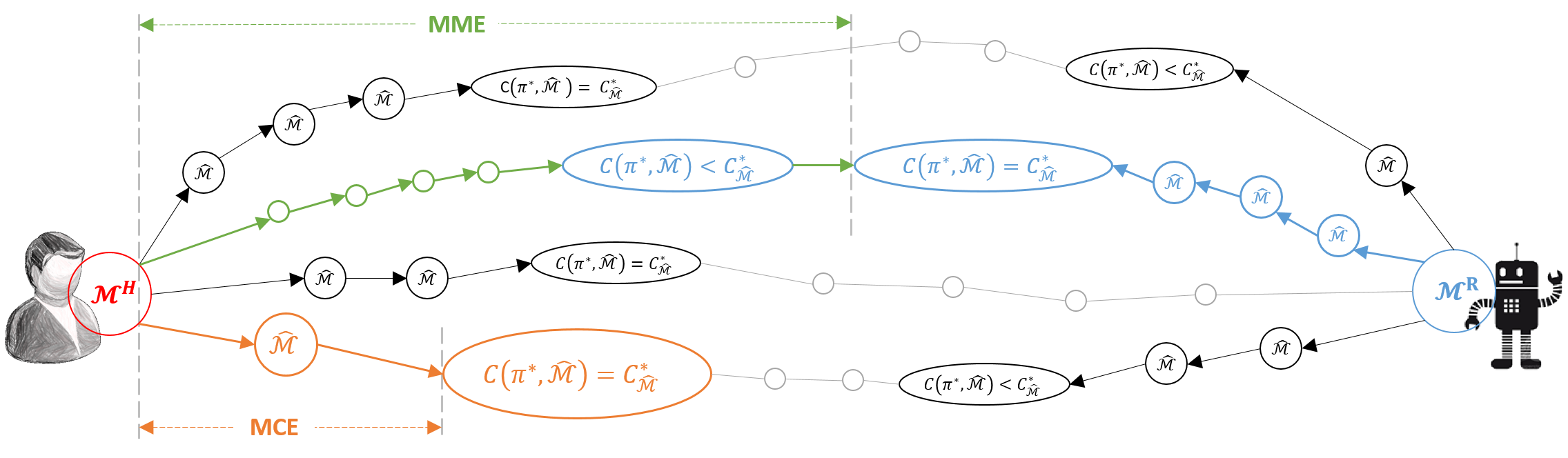}
\caption{Illustration of model space search for MCE \& MME.}
\label{picpic}
\end{figure*}
 
\vspace{-10pt}
\paragraph{A Minimally Monotonic Explanation (MME)} is the shortest  explanation that preserves both completeness and monotonicity, given the MRP 
$\langle \pi^*, \langle \mathcal{M}^R, \mathcal{M}^H\rangle\rangle$, i.e.
\begin{itemize}
\item[] $\mathcal{E}^{MME} = \argmin_{\mathcal{E}}|\Gamma(\widehat{\mathcal{M}}) \Delta \Gamma(\mathcal{M}^H)|$ with R1 \& R3.
\end{itemize}

\vspace{5pt}
The last constraint enforces the monotonicity requirement. This means that beyond the model obtained from the minimally monotonic explanation, there do not exist any models which are not explanations of the same MRP, while at the same time making as few changes to the original problem as possible. It follows that this is the largest set of changes that can be done on the planner's planning problem $\mathcal{M}^R$ and still find a model $\widehat{\mathcal{M}}$ where $C(\pi^*, \widehat{\mathcal{M}}) = C^{*}_{\widehat{\mathcal{M}}}$ - we are going to use this property in the search for MMEs.
 
\vspace{-10pt}
\paragraph{Proposition 3} 
\label{prop1}
$\mathcal{E}^{MME} = \argmax_{\mathcal{E}}|\Gamma(\widehat{\mathcal{M}}) \Delta \Gamma(\mathcal{M}^R)|$ such that 
$\forall\hat{\mathcal{M}}~\Gamma(\hat{\mathcal{M}}) \Delta \Gamma(\mathcal{M}^R) \subseteq \Gamma(\widehat{\mathcal{M}}) \Delta \Gamma(\mathcal{M}^R)$ it is guarantee to have $C(\pi^*, \hat{\mathcal{M}}) = C^{*}_{\hat{\mathcal{M}}}$.
 
\vspace{5pt}
We also note that an MME solution may not be unique to an MRP problem. This can happen when there are multiple model differences supporting the same causal links in the plan - an MME can get by (i.e. guarantee optimality in the modified model) by only exposing one of them to the human.
 
\vspace{-10pt}
\paragraph{Proposition 4} 
\label{prop4}
MMEs are not unique, i.e. there might be multiple minimally monotonic solutions to an MRP.

\vspace{5pt}
We also note that even though MCEs are an abridged version of an MME, it is easy to see that an MCE may not necessarily be part of an actual MME. This is due to the non-uniqueness of MMEs. Thus, we emphasize - 
 
\vspace{-10pt}
\paragraph{Proposition 5} An MCE may not be a subset of an MME, but it is always smaller or equal in size, i.e. $|MCE|\leq|MME|$.

\vspace{-10pt}
\paragraph{Model-space search for MMEs}
 
This is similar to the model-space search for MCEs described before, but this time starting from the robot's model $\mathcal{M}^R$ instead. 
The goal here is to find the largest set of model changes for which the explicability criterion becomes invalid for the first time (due to either suboptimality or inexecutability). This 
requires a search over the entire model space, as described in detail in Algorithm \ref{algo2}. 
We can leverage Proposition \ref{prop1} to reduce our search space. 
Starting from $\mathcal{M}^R$, given a set of model changes $\mathcal{E}$ where $\delta_{\mathcal{M}_R,\mathcal{M}_H}(\Gamma(\mathcal{M}^R), \mathcal{E}) = \Gamma(\widehat{\mathcal{M}})$ and $C(\pi^*,\widehat{\mathcal{M}}) > C^{*}_{\widehat{\mathcal{M}}}$, no superset of $\mathcal{E}$ can lead to an MME solution. 
In Algorithm \ref{algo2}, we keep track of such unhelpful model changes in the list h\_list. 
The variable $\mathcal{E}^{MME}$ keeps track of the current best list of model changes. Whenever we find a new set of model changes where $\pi*$ is optimal and is larger than $\mathcal{E}^{MME}$, we update $\mathcal{E}^{MME}$ with  $\mathcal{E}$. The resulting MME is all the  possible model changes that did not appear in $\mathcal{E}^{MME}$.

Figure \ref{picpic} contrasts MCE with MME search. MCE search starts from $\mathcal{M}^{H}$, computes updates $\widehat{\mathcal{M}}$ towards $\mathcal{M}^{R}$ and returns the first node (indicated in orange) where $C(\pi^*, \widehat{\mathcal{M}}) = C^{*}_{\widehat{\mathcal{M}}}$.
MME search starts from $\mathcal{M}^{R}$ and moves towards $\mathcal{M}^{H}$. 
It finds the longest path (indicated in blue) where $C(\pi^*, \widehat{\mathcal{M}}) = C^{*}_{\widehat{\mathcal{M}}}$ for all $\widehat{\mathcal{M}}$ in the path. The MME (shown in green) is the rest of the path towards $\mathcal{M}^{H}$.

\subsubsection{Approximate Solution for MCEs}

Both MCEs and MMEs may be hard to compute - in the worst case it involves a search over the entire space of model differences. 
Thus the biggest bottleneck here is the check for optimality of a plan given a new model. 
A check for necessary or sufficient conditions for optimality, without actually computing optimal plans can be used as a powerful tool to further prune the search tree. 
In the following section, we thus investigate an approximation to an MCE by employing a few simple proxies to the optimality test.
By doing this we lose the completeness guarantee but improve the computability of an explanation.
Specifically, we replace the equality test in line 12 of Algorithm \ref{algo1} by the following rules -

\begin{enumerate}
\item $\delta_{\widehat{\mathcal{M}}}(\mathcal{\widehat{I}}, \pi^*_R) \models \mathcal{\widehat{G}}$; \textbf{and}
\item $C(\pi^*_R, \widehat{\mathcal{M}}) < C(\pi^*_R, \mathcal{M}^H)$ \textbf{or} $\delta_{\widehat{\mathcal{M}}}(\mathcal{\widehat{I}}, \pi^*_H) \not\models \mathcal{\widehat{G}}$; \textbf{and}
\item Each action contributes at least one causal link to $\pi^*_R$.
\end{enumerate}
 
The first criterion simply ensures that the plan $\pi^*_R$ originally computed by the planner is actually valid in the new hypothesis model. Criterion (2) requires that this plan has either become better in the new model or at least that the human's expected plan $\pi^*_H$ has been disproved. Finally, in Criterion (3), we ensure that for each action $a_i \in \pi^*_R$ there exists an effect $p$ that satisfies the precondition of at least one action $a_k$ (where $a_{i} \prec a_{k}$) and there exists no action $a_j$ (where $a_i \prec a_j \prec a_k)$ such that $ p \in \textit{eff}^-(a_j)$.
 
\vspace{-10pt}
\paragraph{Proposition 6}
Criterion (3) is a necessary condition for optimality of $\pi^*$ in $\widehat{\mathcal{M}}$.
 
\vspace{-10pt}
\paragraph{Proof}
Assume that for an optimal plan $\pi^*_R$, there exists an action $a_i$ where criterion (3) is not met. 
Now we can rewrite $\pi^*_R$ as $\pi'_R= \langle a_0, a_1, \ldots,a_{i-1},a_{i},a_{i+1},\ldots, a_n, a_{n+1} \rangle$, where $\textit{pre}(a_0) = \phi$ and $\textit{eff}^+(a_0) = \{\mathcal{I}\}$ and $\textit{pre}(a_{n+1}) = \{\mathcal{G}\}$ and $\textit{eff}(a_{n+1}) = \phi$. It is easy to see that $\delta_{\widehat{\mathcal{M}}}(\phi, \pi'_R)\models \mathcal{G}$. 
Now let us consider a cheaper plan $\hat{\pi'_R}= \langle a_0, a_1, \ldots,a_{i-1},a_{i+1},\ldots, a_n, a_{n+1} \rangle$. Since $a_i$ does not contribute any causal links to the original plan $\pi^*_R$, we will also have $\delta{\widehat{\mathcal{M}}}(\phi, \hat{\pi}'_R) \models \mathcal{G} $. This contradicts our original assumption of $\pi^*_R$ being optimal, hence proved.

\begin{table*}
\tiny
\centering
  \begin{tabular}{r|c|c|c|c|c|c|c|c|c|c|c|c|c}
    \toprule
    \multirow{3}{*}{Domain Name} &
    \multirow{3}{*}{Problem} &
     \multicolumn{2}{c|}{MPE} &
     \multicolumn{2}{c|}{PPE} &
      \multicolumn{2}{c|}{MME} &
      \multicolumn{2}{c|}{MCE (exact} &
      \multicolumn{2}{c|}{MCE (exact} &
      \multicolumn{2}{c}{MCE} \\
      &
      & \multicolumn{2}{c|}{(ground truth)}
      & \multicolumn{2}{c|}{} 
      & \multicolumn{2}{c|}{(exact)} 
      & \multicolumn{2}{c|}{w/o heuristic)}
& \multicolumn{2}{c|}{with heuristic)} 
      & \multicolumn{2}{c}{(approximate)} 
      \\[1ex] \cline{3-13}
      & 
	      & {size} & {time}& {size} & {time} & {size} & {time} & {size} 
      & {time} & {size} & {time} & {size} & {time} \\
      \midrule
      \midrule
    \multirow{4}{*}{BlocksWorld} &1& \multirow{4}{*}{10} & \multirow{4}{*}{n/a}  & 5 & \multirow{4}{*}{n/a} & 3   &  1100.8   &  2 & 34.7 & 2 & 18.9&2  & 19.8\\ 
     & 2 & & & 8 &&4  & 585.9 &  3 & 178.4& 3 & 126.6 & 3  & 118.8  \\
        & 3 & &  &4&& 5  & 305.3  &  2 & 34.7&2 & 11.7 &2 & 11.7  \\
       & 4 & & &7&& 5  & 308.6  & 3  & 168.3& 3 &  73.3& 3  & 73.0  \\
      \midrule
       \multirow{4}{*}{Rover} & 1& \multirow{4}{*}{10} & \multirow{4}{*}{n/a}  &10&\multirow{4}{*}{n/a}&  2 & 2093.2 &   2  & 111.3 &2 & 100.9 & 2  & 101.0 \\
       &2& &&  10&&2 & 2018.4 &  2  & 108.6& 2 & 101.7  & 2  & 102.7  \\
       &3& &&  10&&2 & 2102.4&   2 &  104.4 &2  & 104.9 &  2 &  102.5  \\
        &4& &&  9&&1 & 3801.3&   1 & 13.5 &1 & 12.8  &   1 & 12.5   \\
          \midrule
\multirow{4}{*}{Logistics} &1& \multirow{4}{*}{5}  & \multirow{4}{*}{n/a} & 5&\multirow{4}{*}{n/a}&4  &  13.7 & 4  & 73.2& 4  & 73.5 & 4 &  63.6  \\
& 2&&& 5&&4  & 13.5  & 4  & 73.5& 4  & 71.4  & 4  & 63.3 \\
        & 3&&& 5&& 5  & 8.6 & 5  &  97.9& 5  & 100.4 & 3  & 36.4  \\
       &4& &&  5&&5 &  8.7 &  5 &  99.2& 5  & 95.4 &  3 &  36.4  \\
    \bottomrule
    \bottomrule
  \end{tabular}
\caption{
Comparison of MCEs and MMEs
}
\label{tab1}
\end{table*}
 
\begin{table}[tbp!]
\tiny
\centering
\begin{tabular}{c|cccc}
\toprule
$|\mathcal{M}^R\Delta\mathcal{M}^H|$ & problem-1 & problem-2 & problem-3 & problem-4 \\ 
\midrule
\midrule
3 & 2.2          & 18.2          & 4.7          & 18.5          \\ 
5 & 6.0     &   109.4        & 15.4         & 110.2          \\ 
7 & 7.3          & 600.1          & 23.3          & 606.8          \\ 
10 & 48.4          & 6849.9         & 264.2         & 6803.6         \\ 
\bottomrule
\bottomrule
\end{tabular}
\caption{MCE search time for increasing model differences.}
\label{tab2}
\end{table}

\begin{table}[tbp!]
\tiny
\centering
\begin{tabular}{@{}c|cccc@{}}
\toprule
BlocksWorld & problem-1 & problem-2 & problem-3 & problem-4 \\ \midrule\midrule
\begin{tabular}[c]{@{}c@{}}Number of nodes expanded\\ for MME (out of $1024$)\end{tabular} & 128   &       64   &    32      &   32      \\ 
\bottomrule
\bottomrule
\end{tabular}
\caption{Usefulness of Proposition 3 in pruning MME search.}
\label{tab3}
\end{table}
 
\section{Empirical Evaluations}
 
Our explanation generation system (as previewed in the Fetch domain) integrates calls to Fast-Downward \cite{helmert2006fast} for planning, VAL \cite{howey2004val} for plan validation, and pyperplan \cite{pyperplan} for parsing. 
The results reported here are from experiments run on a 12 core Intel(R) Xeon(R) CPU with an E5-2643 v3@3.40GHz processor and a 64G RAM.
%The latest version of the code will be available at \url{https://goo.gl/Bybq7E}.
 The latest version of the code will be available at https://goo.gl/Bybq7E.
We use three planning domains - BlocksWorld, Logistics and Rover - for our experiments. 
In order to generate explanations we created the human model by randomly removing parts (preconditions and effects) of the action model.
Though the following experiments are only pertaining to action model differences, it does not make any difference at all to the approaches, given the way the state was defined. 
Also note that these removals, as well as the corresponding model space search, was done in the lifted representation of the domain.
 
\vspace{-10pt}
\paragraph{Table \ref{tab1}}
In Table \ref{tab1} we make changes at random to the domains and measure the number of explanations produced and the time taken (in secs) to produce them, against the ground truth.
Observe the gains produced by the heuristic in terms of time spent on each problem. Further, note how close the approximate version of MCEs are to the exact solutions. 
As expected, MME search is significantly costlier to compute than MCE. However, note that both MCEs and MMEs are \emph{significantly smaller} in size ($\sim 20\%$) than the total model difference (which can be arbitrarily large) in certain domains, further underlining the usefulness of generating minimally complete explanations as opposed to dumping the entire model difference on the human. A general rule of thumb is -
\begin{equation*}
\small
|~\text{approx.}~MCE~| \leq |~\text{exact.}~MCE~| < |~MME~| << |MPE| 
\end{equation*}
 
Note that the time required to calculate an MME in the Logistics problems is lower than that for the corresponding MCE. 
This is because for most of these problems a single change in the planner's model made the plan be no longer optimal so that the search ended after checking all possible unit changes. In general, closer an MCE is to the total number of changes shorter the MME search would be.
Also note how PPE solutions, though much easier to compute, do not have completeness and monotonicity properties, and yet often spans the entire model difference, containing information that are not needed to support the optimality of the given plan.
 
\vspace{-10pt}
\paragraph{Table \ref{tab2}} We now increase the number of changes in the human model in BlocksWorld, and illustrate the relative time (in secs) taken to search for exact MCEs in Table \ref{tab2}. As expected there is an exponential increase in the time taken, which can be problematic with even a modest number of model differences. This further highlights the importance of finding useful approximations to the explanation generation problem.
 
\vspace{-10pt}
\paragraph{Table \ref{tab3}} Finally, we demonstrate how Proposition 3
reduces the number of nodes that need to be searched to find MMEs in random problems from the BlocksWorld domain with 10 faults in the human model, as opposed to the total possible $2^{10}$ models that can be evaluated - equal to the cardinality of the power set of model changes $|\mathcal{P}(\Gamma(\mathcal{M}^R)\Delta\Gamma(\mathcal{M}^H))|$. 
 
\section{Conclusions and Future Work}

In this paper, we argued that to explain its plans to the human agents in the loop, an AI system needs to explicitly acknowledge that the human may be using a different model than it does. Explanations in this multi-model setting become a process of identifying and reconciling the relevant differences between the models. 
One immediate future direction is to allow for human's models that are of different form and/or level of abstraction than the robot's, so as to allow effective learning of the human's models (c.f. \cite{tian2016discovering,exp-yu}), as well as allow for different cognitive abilities of the human using $\epsilon$-optimality or top-K plans \cite{RiaSohUdrSPARK14} for hypothesis generation. 

In cases where the ground truth is not known, the explanation process might also need to consider distributions over relevant models, and iterative refinement of the same via dialog with the human. 
Work on plan monitoring \cite{fritz2007monitoring} can also provide clues to speeding up the search process by providing proxies to the optimality check.

Also note that we insisted that explanations
must be compatible with the planner’s model.
If this requirement is relaxed, it allows the planner 
to generate \emph{alternative explanations} 
that it knows are not true, and thus deceive the human. 
While endowing the planner with such abilities
may warrant significant ethical concerns, we note that the
notion of white lies, and especially the relationship between
explanations, excuses and lies \cite{Boella2009}
has received very little attention \cite{vanDitmarsch2014} 
and affords a rich set of exciting research problems.

%\vspace{-10pt}
\section*{Acknowledgments} This research is supported in part by the ONR grants N00014161-2892, N00014-13-1-0176, N00014- 13-1-0519, N00014-15-1-2027, and the NASA grant NNX17AD06G.

\cleardoublepage
\bibliographystyle{named}
\bibliography{curr_bib}
 
\end{document}